\documentclass{article}

\usepackage{PRIMEarxiv}

\usepackage[utf8]{inputenc} 
\usepackage[T1]{fontenc}    
\usepackage{hyperref}       
\usepackage{url}            
\usepackage{booktabs}       
\usepackage{amsfonts}       
\usepackage{nicefrac}       
\usepackage{microtype}      
\usepackage{lipsum}
\usepackage{graphicx}
\usepackage{makecell}
\usepackage{subfigure}
\graphicspath{{media/}}     

\title{DPANET:Dual Pooling Attention Network for Semantic Segmentation
}

\author{
  Dongwei Sun, Zhuolin Gao \\
  Northwestern Polytechnical University \\
  Xi'an\\
  \texttt{\{sundongwei\}@outlook.com} \\
}

\begin{document}
\maketitle

\begin{abstract}
Image segmentation is a historic and significant computer vision task. With the help of deep learning techniques, image semantic segmentation has made great progresses. Over recent years, based on guidance of attention mechanism compared with CNN which overcomes  the problems of lacking of interaction between different channels, and effective capturing and aggregating contextual information. However, the massive operations generated by the attention mechanism lead to its extremely high complexity and high demand for GPU memory. For this purpose, we propose a lightweight and flexible neural network named Dual Pool Attention Network(DPANet).
The most important is that all modules in DPANet generate \textbf{0} parameters. The first component is spatial pool attention module, we formulate an easy and powerful method densely to extract contextual characteristics and reduce the amount of calculation and complexity dramatically.Meanwhile, it demonstrates the power of even and large kernel size. The second component is channel pool attention module. It is known that the computation process of CNN incorporates the information of spatial and channel dimensions. So, the aim of this module is stripping them out, in order to construct relationship of all channels and heighten different channels semantic information selectively. Moreover, we experiments on segmentation datasets, which shows our method simple and effective with low parameters and calculation complexity. \footnote{The paper was written in 2020}
\end{abstract}


\section{Introduction}
Humans have limited visual attention, so we pay close attention to most essential parts, others will have a low-level attention. The core of the attention mechanism we created for computer vision is for the neural system networks to mimic human visual attention and learn to autonomously focus on the focal information and ignore other non-essential information. In a word, the core of attention mechanism is to focus on highlight information parts but whole system. 

Recent years, there is growing works \cite{vaswani2017attention} \cite{mnih2014recurrent} on combining such attention mechanisms with deep learning, where main principle is to form new weighted maps in a way that identifies key features. There are two branches in attention field, channel attention and spatial attention. SENet\cite{senet} and Non-local \cite{wang2018non} are representative of their respective fields.
\begin{figure}[ht]
  \centering
   \includegraphics[width=0.6\linewidth]{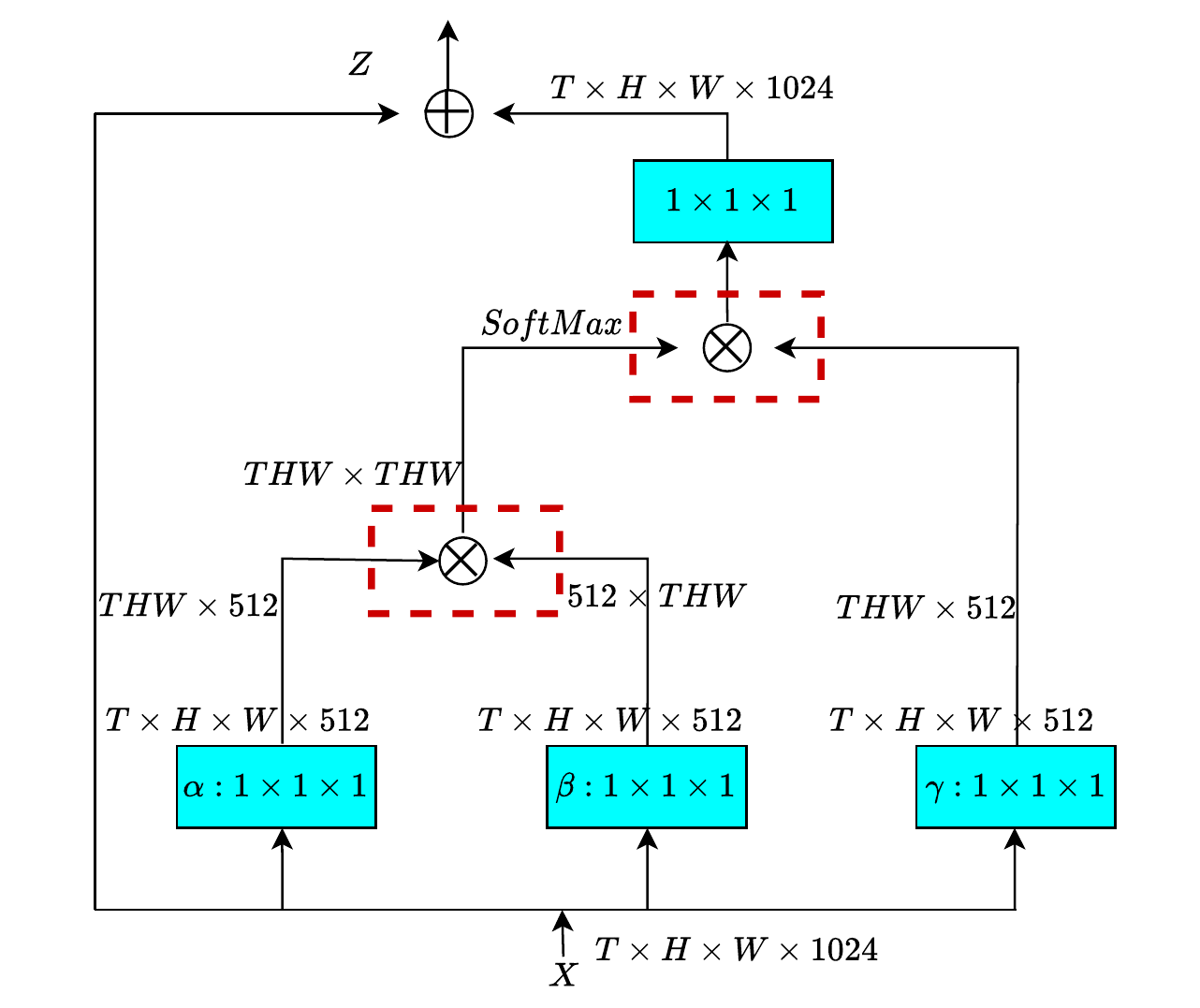}
    \caption{Non-local}
   \label{non-local}
\end{figure}
However, there are shortcomings in their works. For example, the work for non-local is missing the information measure in the channel dimension, and the computational complexity of the that is very high, main calculations on red blocks as shown in Figure \ref{non-local}  especially when the feature map is large. Secondly, for SENet, which has a poor versatility and coarsely reconstruct correlation of channels.

To alleviate these problems in an efficient way with sacrificing low precision is purpose of our DPANet. We propose spatial pool attention module and channel pool attention module. The former inspired by pyramid operations \cite{zhao2017pyramid} \cite{chen2017deeplab} \cite{chen2017rethinking} and large, even kernels \cite{Peng_2017_CVPR} \cite{wu2019convolution}. In process of using pyramid pooling, not only is the amount of operations significantly reduced, but also the addition of even size pooling layers compared to traditional odd numbers pooling layers allow for intensive and seamless features extracted. Therefore, we designed pyramid pool module with odd and even different sizes to decrease non-local complexity dramatically. As for channel attention, although variants of SENet continue to improve the former's work in the channel dimension, they are still accompanied by the generation of parameters and the lack of mutual metrics between channels. So, we improve coarser method in SENet series by pooling operation on each channel and measure the differences on each other with no parameters.

In summary, our contributions are as follows:
\begin{enumerate}
    \item We propose spatial pool attention module embedded pool unit, which is an easy and efficient way to generate attention map with no params.
    \item We demonstrate that re-designing pyramid pool mode by inserting even size is efficaciously.
    \item We show new channel pool attention module that fast and simply to build all channel information.
    \item We showcase that our DPANet outperforms results on segmentation datasets.
\end{enumerate}





\section{Related Work}
In this section we will review recently works include two parts: 1) semantic segmentation; 2) attention module.

\textbf{Semantic segmentation.} FCN \cite{long2015fully} as a pioneer of semantic segmentation using deep learning has greatly influence the subsequent works. The most important is DeepLab series \cite{chen2017deeplab} \cite{chen2017rethinking} \cite{chen2018encoder}, which invent dilated kernel for getting larger respective field and atrous spatial Pyramid Pooling(ASPP) module. PSPNet \cite{zhao2017pyramid} and EncNet \cite{zhang2018context} in order to obtain global contextual information by pyramid pool module or context encoding module.RefineNet \cite{lin2017refinenet}, a generic multi-path network that explicitly utilizes all the information available throughout the downsampling process, using remote residual connections to achieve high-resolution predictions.G-FRNet \cite{amirul2017gated} performs coarse predictions and then progressively refines the details by efficiently integrating local and global contextual information in the refinement phase, introducing gate units that control the forward transmission of information to filter ambiguities.

\textbf{Attention module.} Attention mechanism \cite{vaswani2017attention} has been proposed in recent year. Attention algorithms have been adopted with high speed in the field of artificial intelligence, not only in NLP, text translation, but also in computer vision and image process. At the same time, more and more self-attentive mechanisms have been introduced into the image processing field. SENet \cite{senet} enhances the expressiveness of the network by modeling the image channels with an attention mechanism. Chen \cite{chen2017rethinking} use the generation of several attention masks to fuse or predict the feature maps of different branches. wang's \cite{wang2018non} non-local was the first to find that CNNs are a local operation and can only act on local regions. Therefore, a guiding matrix that reweights the information of the original image by computing the relationship between each pixel point in the spatial of image, which brings great benefits for semantic segmentation using long-range dependencies. OCNet \cite{yuan2018ocnet} and DANet \cite{fu2019dual} use self-attention to obtain contextual information. PSA \cite{zhao2018psanet} proposes to learn an attention map for adaptive and targeted aggregating contextual information.
\section{Method}
In this section, we explain our Dual Pooling Attention Network (DPANet) amply. Step one, we give an overview of the framework,then elaborate the details of each module and in the end how the framework to be combined.
\subsection{Overview}
\begin{figure}[ht]
  \centering
   \includegraphics[width=\linewidth]{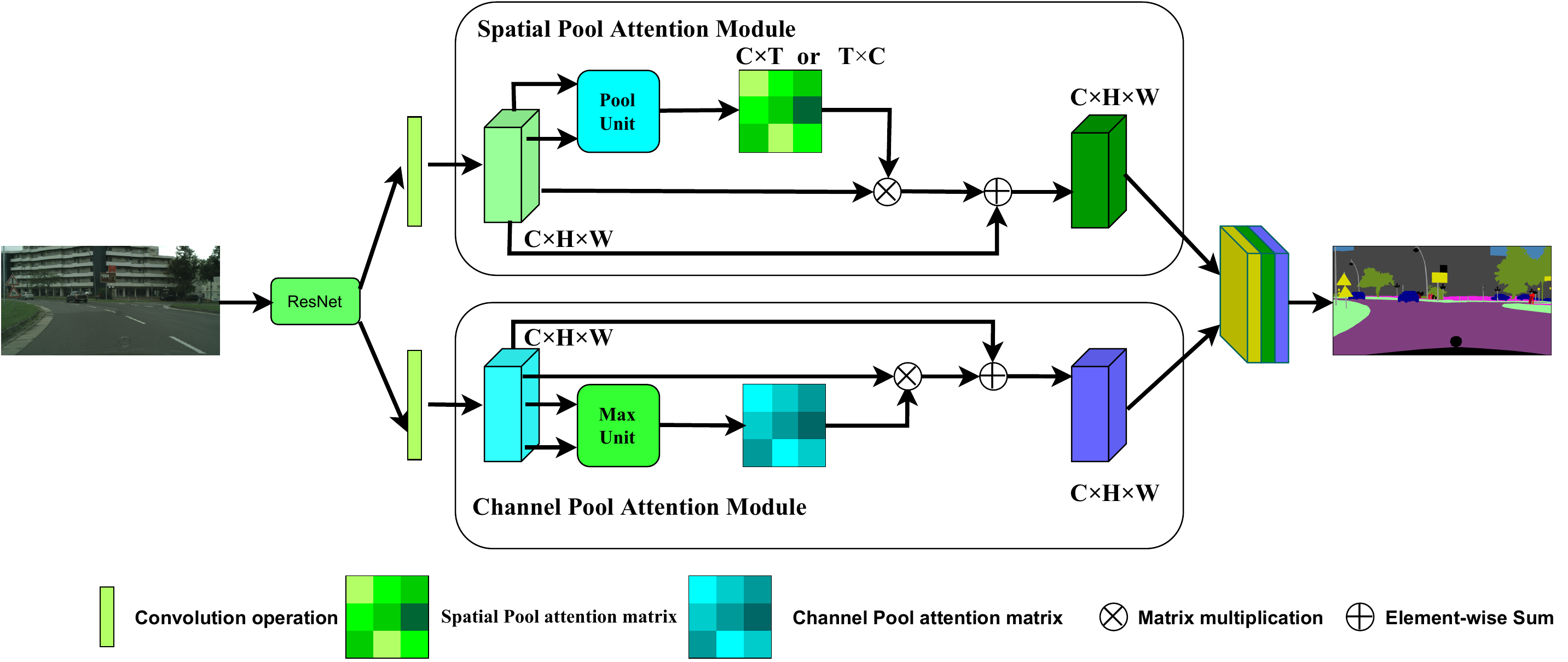}
    \caption{Overview of Dual Pooling Attention Network.There are two branches in the framework:the upper part is spatial attention module, the lower one is channel attention module.Their outputs to be concatenated is final output.}
   \label{dpanet}
\end{figure}

DPANet contains two parts: spatial attention module and channel attention module as illustrated Figure \ref{dpanet}.We adopt ResNet series as backbone.Then,the output of ResNet is feature map $X\in \mathbb{R}^{C\times H\times W}$,where $H$ is the height of the feature map and $W$ is the width of the feture map,and $C$ is the channel dimension.The $X$ is fed into two ways:spatial pooling attention module(SPA) and channel pooling attention module(CPA).For the first step is SPA module,we designed Pool Unit on $X$ to generate weight-attention mask for propose of aggregating long-range and whole feature map contextual information with a low operations and parameters calculations.Next,the second way is CPA,which is designed for stripping spatial and channel's mixed features out,getting a rebalance by filtering the salient information on each channels' feature map.
\subsection{Spatial Pooling module}
The classic attention model is Non-local Neural Networks\cite{wang2018non}, the pipeline of that is: taking a feature map $X\in \mathbb{R}^{C\times H\times W}$as input,where $C,H,W$ represent the channels,height,width respectively. After that,passed through $1\times 1$ convolutional operations as
\begin{equation}
    \label{eq1}
    \alpha = W_{\alpha}(X), \beta = W_{\alpha}(X), \gamma = W_{\alpha}(X)
\end{equation}
the result is $\alpha=\mathbb{R}^{\hat{C}\times H\times W},\beta=\mathbb{R}^{\hat{C}\times H\times W},\gamma=\mathbb{R}^{\hat{C}\times H\times W}$ for $\hat{C}$ is new numbers of channel. For next will get weighted-attention tensor $M\in \mathbb{R}^{N\times N}$ by
\begin{equation}
    \label{eq2}
    M = \mathrm{SoftMax}(M^{'}),\quad  M^{'}= \alpha^T \times \beta
\end{equation}
where superscript $T$,$N=H\times W$ indicate transpose and total pixels in spatial dimension.The output of attention layer is 
\begin{equation}
    \label{eq3}
    O = M\times \gamma^{T}
\end{equation}

Non-local block is not only a pioneer of attention mechanism but also solving problem for long-range information dependence. However,there is a fatal trouble compared CNN, Pool,etc. is attention mechanism operations require huge GPU memory and tremendous computation, which time complexity is $\mathcal{O}(\hat{C}N^2)=\mathcal{O}(\hat{C}H^2W^2)$.

With the above analysis, main calculations focus on equation \ref{eq2} and \ref{eq3}. So,effective approach is finding a way to replace large number $N$ by small $T$ as
\begin{equation}
    \mathbb{R}^{\hat{C}\times N} \rightarrow \mathbb{R}^{\hat{C}\times T}
\end{equation}

\begin{figure}[thb] \centering
    \includegraphics[width=0.4\linewidth]{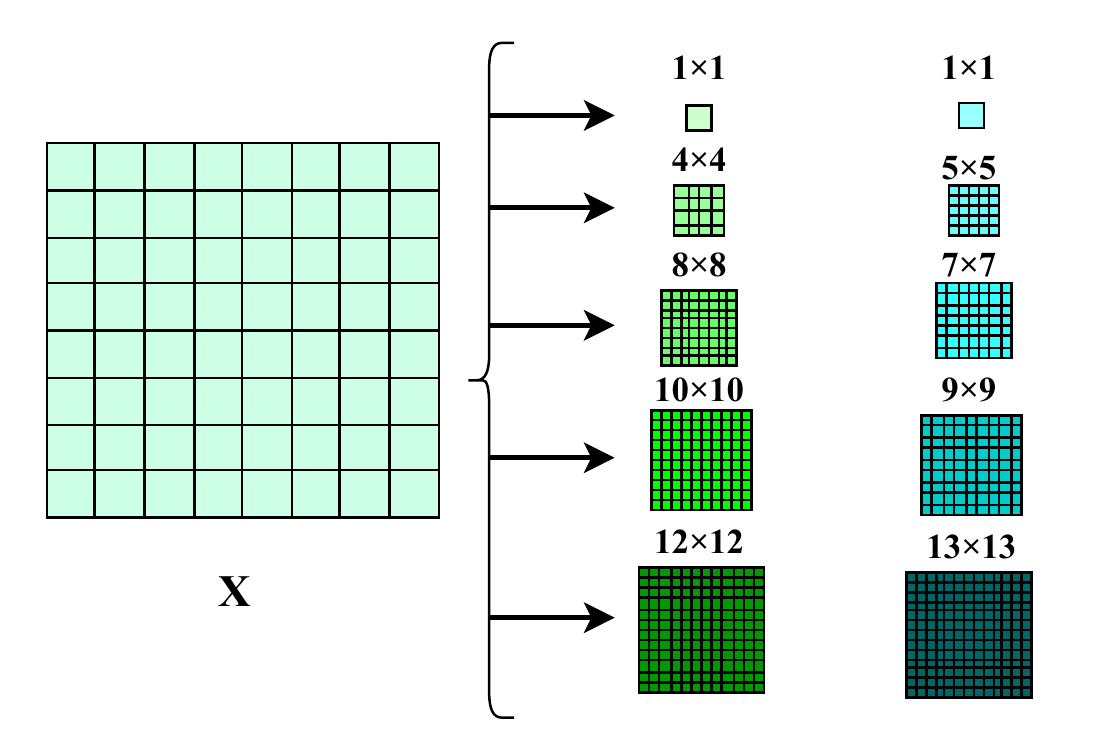}
    \includegraphics[width=0.3\linewidth]{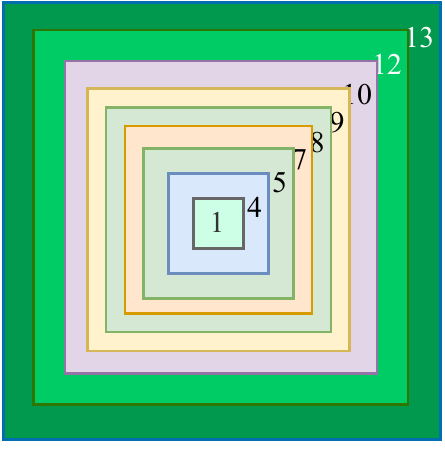}
    \caption{Different scales Pyramid Pool} 
    \label{pool unit}
\end{figure}

Sampling points sparsely on $\alpha$ and $\beta$ by operation $\mathcal{S}$, as a formulation
\begin{equation}
    \alpha_s = \mathcal{S}(\alpha), \beta_s = \mathcal{S}(\beta)
\end{equation}
As the same as before, we get new weighted-attention $\mathcal{M}\in \mathbb{R}^{N\times T}$, and output is
\begin{equation}
    O_s = \mathcal{M}\times \gamma^{T}
\end{equation}
after such an asymmetric matrix transformation, the complexity $\mathcal{O}(\hat{C}NT)$ is  much lower than original $\mathcal{O}(\hat{C}N^2)$.

For alleviate this problem, inspired by PSPNet\cite{zhao2017pyramid} and Deeplab series\cite{chen2017deeplab}\cite{chen2017rethinking}\cite{chen2018encoder}, the proposed pyramid pooling module and ASPP module both capture or fuse multi-scale information with fewer parameters. We proposed a special \textbf{Pool Unit} as shown in Figure \ref{pool unit}.

The advantage of pyramid pool is that could be paralleled generate multi scales pool layers with abundant features but no params produced. The reason for using different sizes of pooling sampling is that, as shown in Figure \ref{pool unit}, the different sizes can cover the entire feature map almost seamlessly and can capture features at different scales, while if a single pooling method is used, the sampling points in the feature map will be omitted, thus affecting the segmentation expression. In our method, $n$ is different sizes referred to $n\in \{1,4,8,10,12\}$ and $n\in \{1,5,7,9,13 \}$, finally total anchor points are 
\begin{equation}
    T = 325 = \sum_{n \in \{1,4,8,10,12\}} n^2 = \sum_{n \in \{1,5,7,9,13\}} n^2
\end{equation}
given input feature map with height and width are both 96, this approach saves nearly $\frac{96\times 96}{325} \approx 29$ times.

\begin{figure}[ht]
  \centering
   \includegraphics[width=0.8\linewidth]{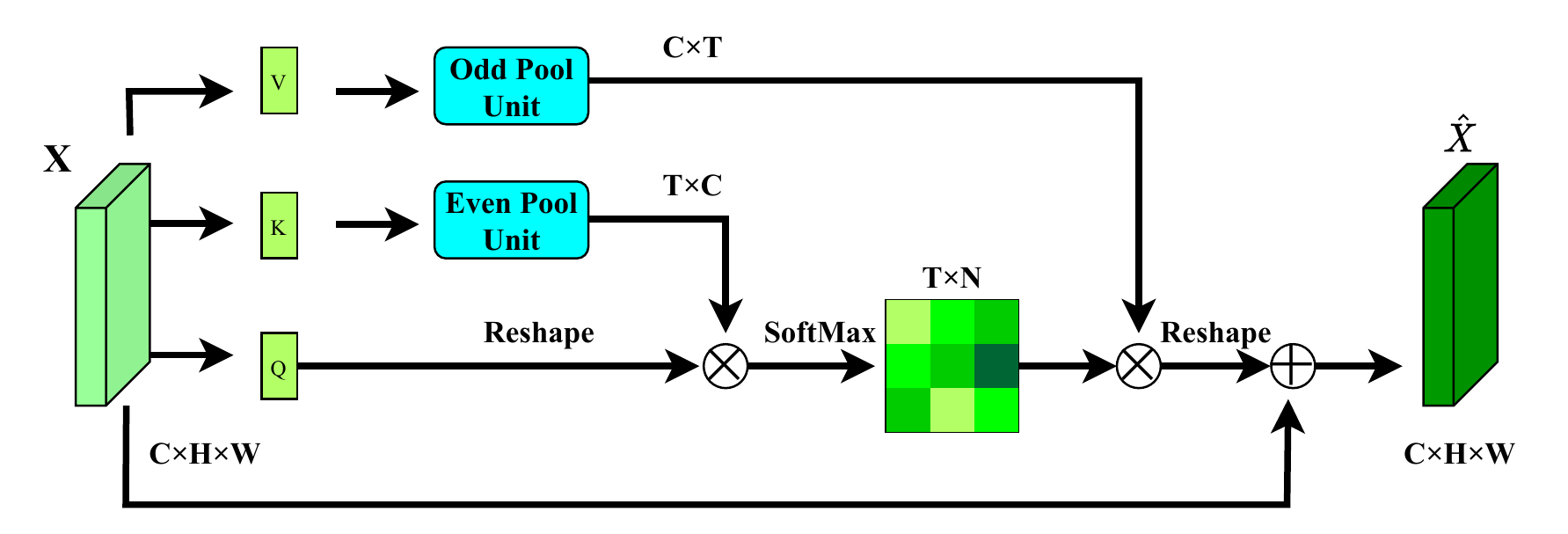}
    \caption{Spatial Pool Module}
   \label{spatial module}
\end{figure}

At last, Pool Unit embedded into spatial attention module as indicated Figure \ref{spatial module},
taking an example as feature map $X$ generate $\{Q,K,V\}\in \mathbb{R}^{C\times H\times W}$ by CNN, where $K$ processed by even Pool unit transposed an asymmetric tensor with shape $\mathbb{R}^{T\times C}$, point in time, applying a softmax to the output of $K$ and $Q$ multiplied as spatial attention map $\mathcal{M}\in \mathbb{R}^{T\times N}$:
\begin{equation}
    \mathcal{M}_{ij} = \frac{exp(K_i \cdot Q_j)}{\sum_{i=1}^N exp(K_i \cdot Q_j)}
\end{equation}
for $\mathcal{M}_{ij}$ is used to measure the impact of $i$th position on $j$th position, the more similarity of location features are, the higher the correlation between them. Meanwhile, the $V$ processed by odd pool unit transposed with shape $\mathbb{R}^{C\times T}$, at the last, we multiply the result with a weight-param $\lambda$ and perform element-wise sum with $X_j$ to obtain final output
\begin{equation}
    \label{spatial output}
    \hat{X}_j = \lambda \sum_{i=1}^N(\mathcal{M}_{ij}\cdot V_i) + X_j
\end{equation}
where $\lambda$ initialized as 0, increasing during training.

Drawing the conclusion from equation \ref{spatial output}, the final information at all positions on the output feature $\hat{X}$ is the result of re-weighting and aggregating with the original features. It is designed to efficiently and powerfully capture the global information with very low computational complexity through spatial pooling attention operations.
\subsection{Channel Pooling module}
Channel as an attribution of image, which illustrates a response of typical object in this dimension, because of the inherent high intensity among channels. That is why we introduce channel attention mechanism to cipher the mapping relationship.

\begin{figure}[ht]
  \centering
   \includegraphics[width=0.8\linewidth]{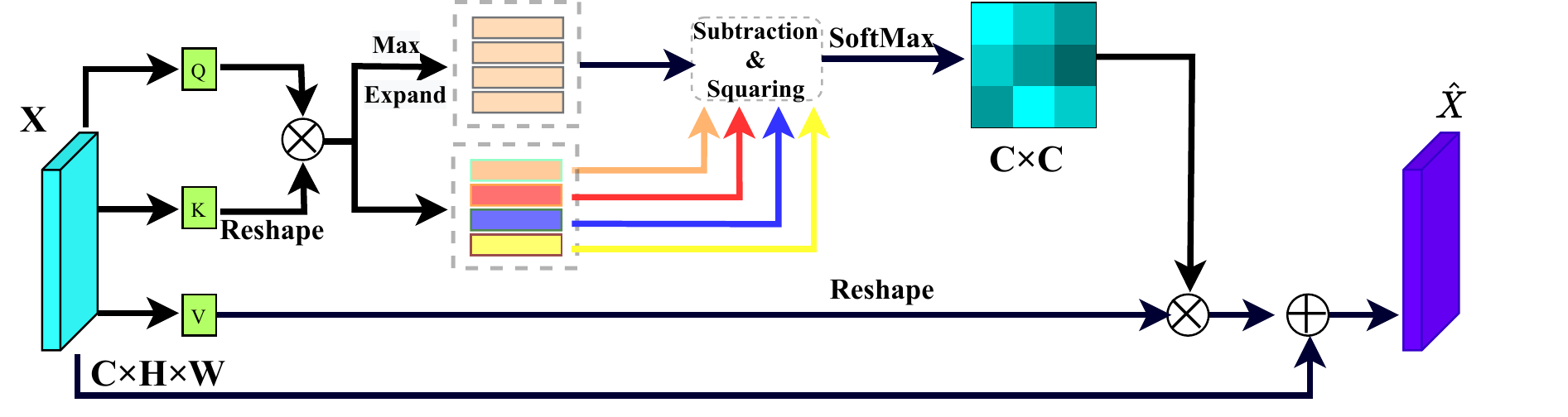}
    \caption{Channel Pool Attention Module}
   \label{channel module}
\end{figure}

We designed channel attention module as Figure \ref{channel module}. Likely operations in spatial attention, we generate $\{Q,K,V\}\in \mathbb{R}^{C\times H\times W}$, reshaping and transposing $K$ to $\mathbb{R}^{N\times C}$, than, the tensors $Q$ and $K$ are multiplied to obtain output $D^{C\times C}$ as a symmetric tensor formulated 
\begin{equation}
    D^{C\times C} = Q^{C\times N}\times K^{N\times C}
\end{equation}

Next, we take max pooling on tensor $D^{C\times C}$ alone channel axis, to get $\hat{D}^{1\times 1\times C}$:
\begin{equation}
    \hat{D} = \mathrm{MaxPool}(D^{C\times C})
\end{equation}
the global distribution of responses on each feature channel can be obtained by this operation. The advantage of using max pooling is that it has more non-linearity and can compensate to a certain extent for the inability of average pooling to distinguish information cues between same or similar objects, which greatly reduces the number of parameters and computations.

For last, we calculate channel attention map by 
\begin{equation}
    \mathcal{C}_j = \hat{D}^{C\times C} - D^{C\times C}\quad\quad \mathrm{OR} \quad\quad
    \mathcal{C}_j = (\hat{D}^{C\times C} - D^{C\times C})^2
\end{equation}
\begin{equation}
    \mathcal{C}_{ji} = \frac{exp(\mathcal{C}_i\cdot \mathcal{C}_j)}{\sum_{i=1}^{N}(\mathcal{C}_i\cdot \mathcal{C}_j)}
\end{equation}
behind this, conveying an idea that is $\mathcal{C}_{ji}$ is determined the impact of channel $i$ to channel $j$, even show diversities of channels. Introducing a param $\mu$ like $\lambda$ in SPA for final output is
\begin{equation}
    \label{cpa output}
    \hat{X}_j = \mu \sum_{i=1}^{C}(\mathcal{C}_{ji}\cdot V_j) + X_j
\end{equation}
where $\mu$ initialized as 0, increasing during training.

The equation \ref{cpa output} shows that the result is composed of two parts. The first part is a re-balancing of the original feature information by channel pooling module for each channel; The second part is the operation of summing the former with the initial input feature map X, that is, the reconstruction of the dependencies and distances between all channels, completely and thoroughly realizing the function and contribution of the channel dimension to image.


\section{Experiments}
We evaluate DPANet on segmentation datasets. Experiments on Cityscapes\cite{Cordts2016Cityscapes}\cite{Cordts2015Cvprw} and PASCAL-VOC2012. The resulets showcase our DPANet outperforms previous works with a simple and effective way on these datasets. In next subsections, we will introduce the details of datasets and experiments details, then we document our experiment results.
\subsection{Experiments Details}
\subsubsection{Datasets}
\textbf{Cityscapes} is one of the most authoritative datasets, which is mainly a segmentation dataset based on environment awareness or scene understanding. It includes most objects in daily life at different times and different places. The dataset is composed of $1024\times 2048$, 5000 images totally.

\textbf{PASCAL VOC 2012} was first constructed, it contained only a few categories:bicycle, car, etc. Now, it is a well-established dataset, the categories are expanded to 20. We mainly uses PASCAL VOC 2012 segmentation dataset, where includes 10,582 images for training, 1,449 images for validation and 1,456 images for testing.
\subsubsection{Implementation Details}
\textbf{Framework} is chosen PyTorch based on PyTorch-Encoding toolkit\cite{zhang2020resnest}.\textbf{Learning rate} followed \cite{Zhang_2018_CVPR} adopted poly learning rate policy where is $lr_{initial}\cdot (1-\frac{iter}{total-iter})^{0.9}$ after each iteration.\textbf{Momentum and weight decay} are set $0.9$ and $0.0001$ respectively. \textbf{Batchsize} are 8 and 16 for Cityscapes and PASCAL VOC respectively. \textbf{Epoches} for Cityscapes are 300, the other is 180. \textbf{Others} are using multi-loss on the final outputs and data augmentation.
\subsection{Experiment Results}
\subsubsection{Results on Cityscapes}
As shown in Table\ref{tab:diff_method}, we took a comparison between non-local block and SPA in following aspects: params memory, FLOPs. We test $96\times 96$($\frac{1}{8}$ of the $768\times 768$) and $256\times 128$($\frac{1}{8}$ of the $2048\times 1024$) as input size on V100 GPU under CUDA10.1.
\begin{table}[thb]\centering
    \caption{Params and FLOPs results on CityScapes}
    \label{tab:diff_method}
    \resizebox{0.48\textwidth}{!}{
    \large
    \begin{tabular}{*{10}{c}}
        \toprule
       Method & Input Size & Params(M) &  FLOPs(G)  \% \\
        \midrule
        NB & $96\times 96$ & 610 & 58.0 \\
        SPA & $96\times 96$ & 0 & 0.0\\
        \bottomrule
        NB & $256\times 128$ & 7797 & 601.4 \\
        SPA & $256\times 128$ & 0 & 0.0 \\
        \bottomrule
    \end{tabular}
    }
\end{table}
The results are obvious, we replace NB block by different scales pooling operations, which generate 0 params and 0 FLOPs. The reason why our method could reduce complexity is that only partly feature points is sampled not whole feature map.To compare the performance with other methods on Cityscapes. Our method attain the competitive performance of 70.1\% as shown in table\ref{t2}.

\begin{table}[ht]
  \caption{Comparison on test set of Cityscapes with other methods.}
  \label{t2}
  \centering
  \begin{tabular}{c|c|c}
    \midrule
    Method& Backbone & MIoU(\%)\\
    \midrule  
    ENet& No& 58.3\\
    ESPNet&No&60.3\\
    CRFasRNN\cite{zheng2015conditional}&FCN&62.5\\
    CGNet\cite{wu2020cgnet}& No& 64.8\\
    LiteSeg-Shufflenet\cite{DBLP:journals/corr/abs-1912-06683}& No& 65.2\\
    DABNet\cite{li2019dabnet}& No& 69.5\\
    DeepLabv2-CRF& ResNet101& 70.0\\
    \midrule
    DPANet&ResNet50&64.2\\
    DPANet&ResNet101&70.1\\ \midrule
  \end{tabular}
\end{table}

\begin{figure}[ht] 
	\centering  
	\subfigure[Images]{
    \begin{minipage}[t]{0.2\linewidth}
      \centering
      \includegraphics[height=2cm,width=.9\linewidth]{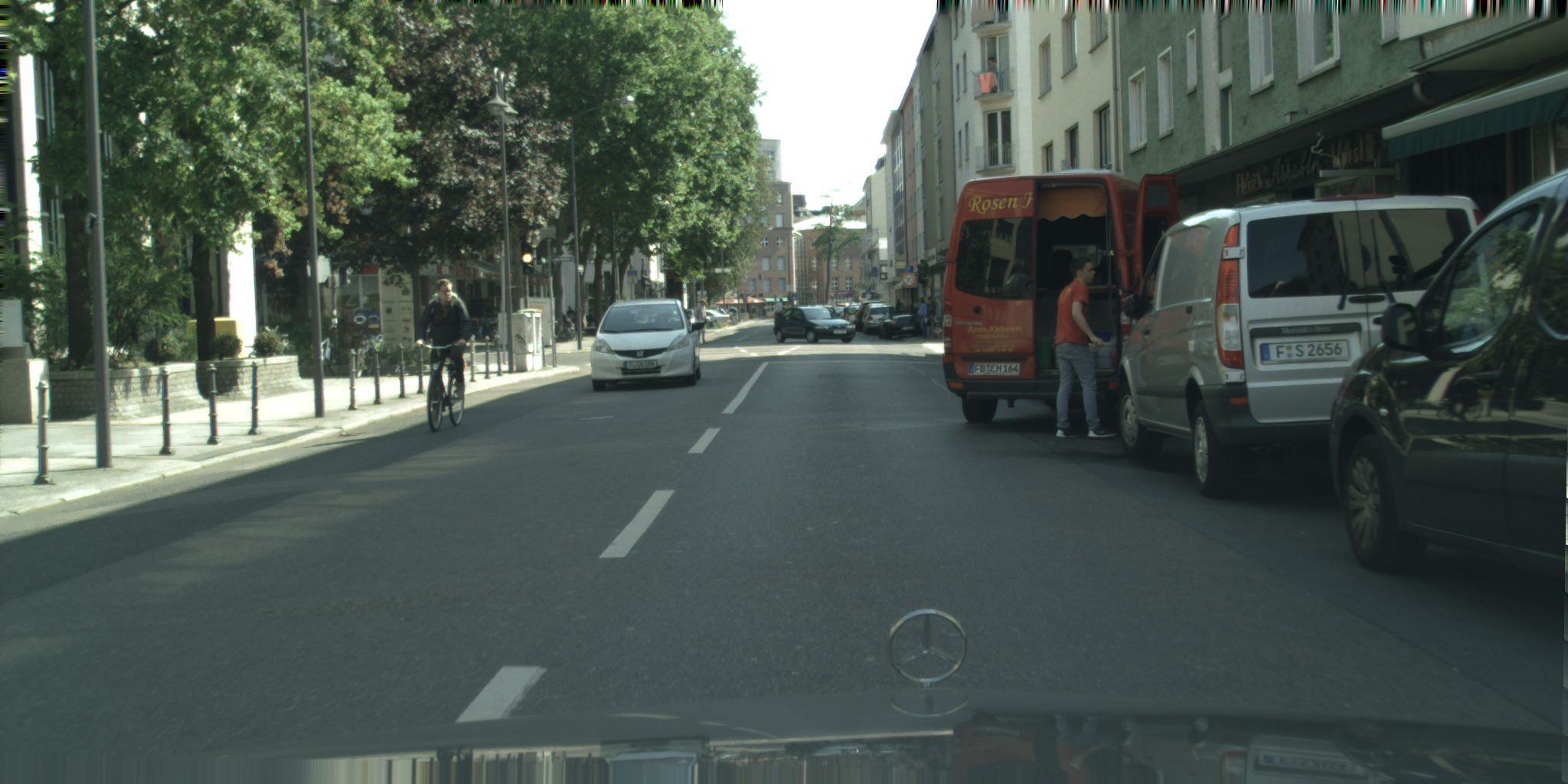}\\
      \vspace{4pt}
      \includegraphics[height=2cm,width=.9\linewidth]{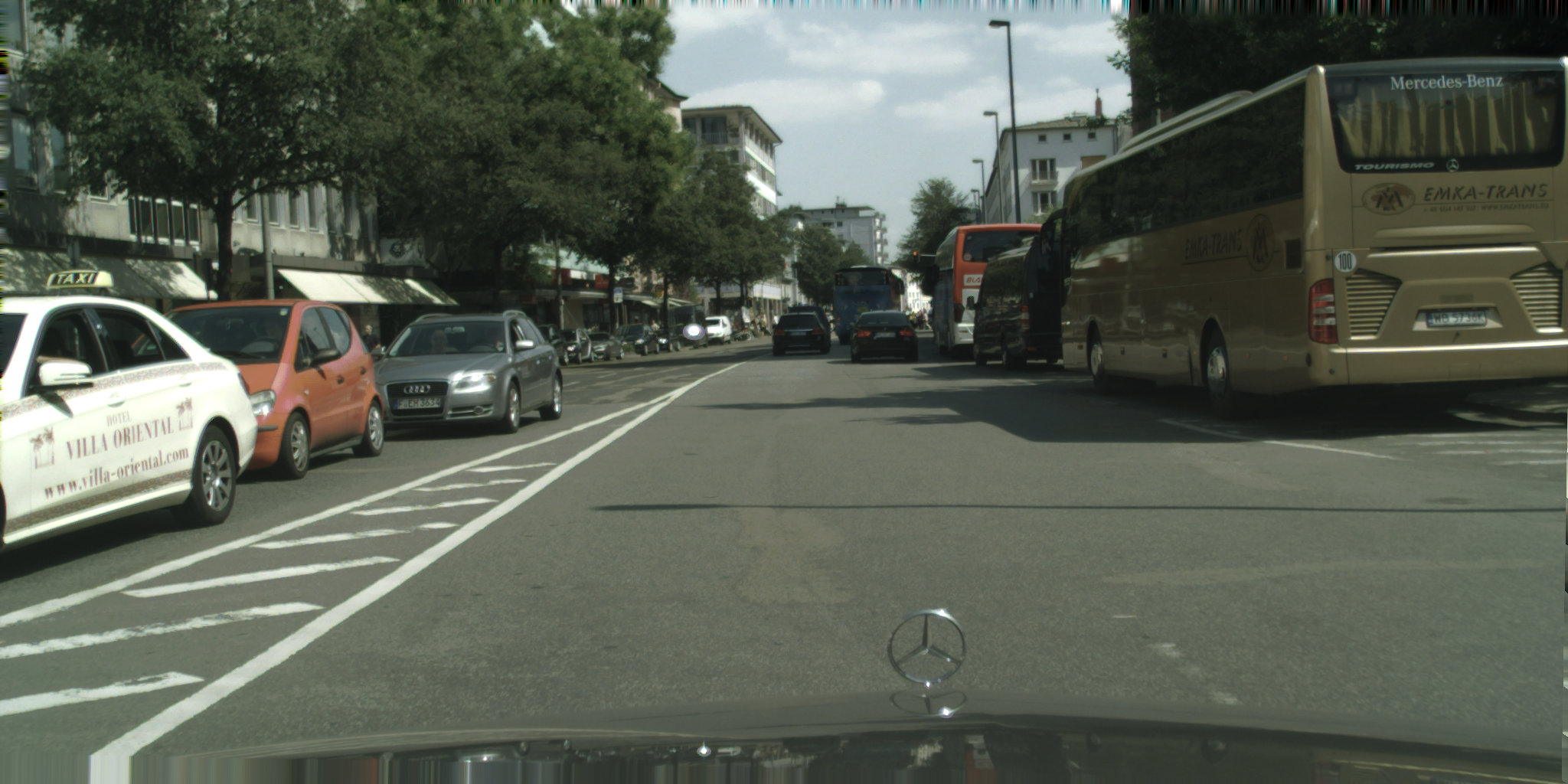}\\
      \vspace{4pt}
      \includegraphics[height=2cm,width=.9\linewidth]{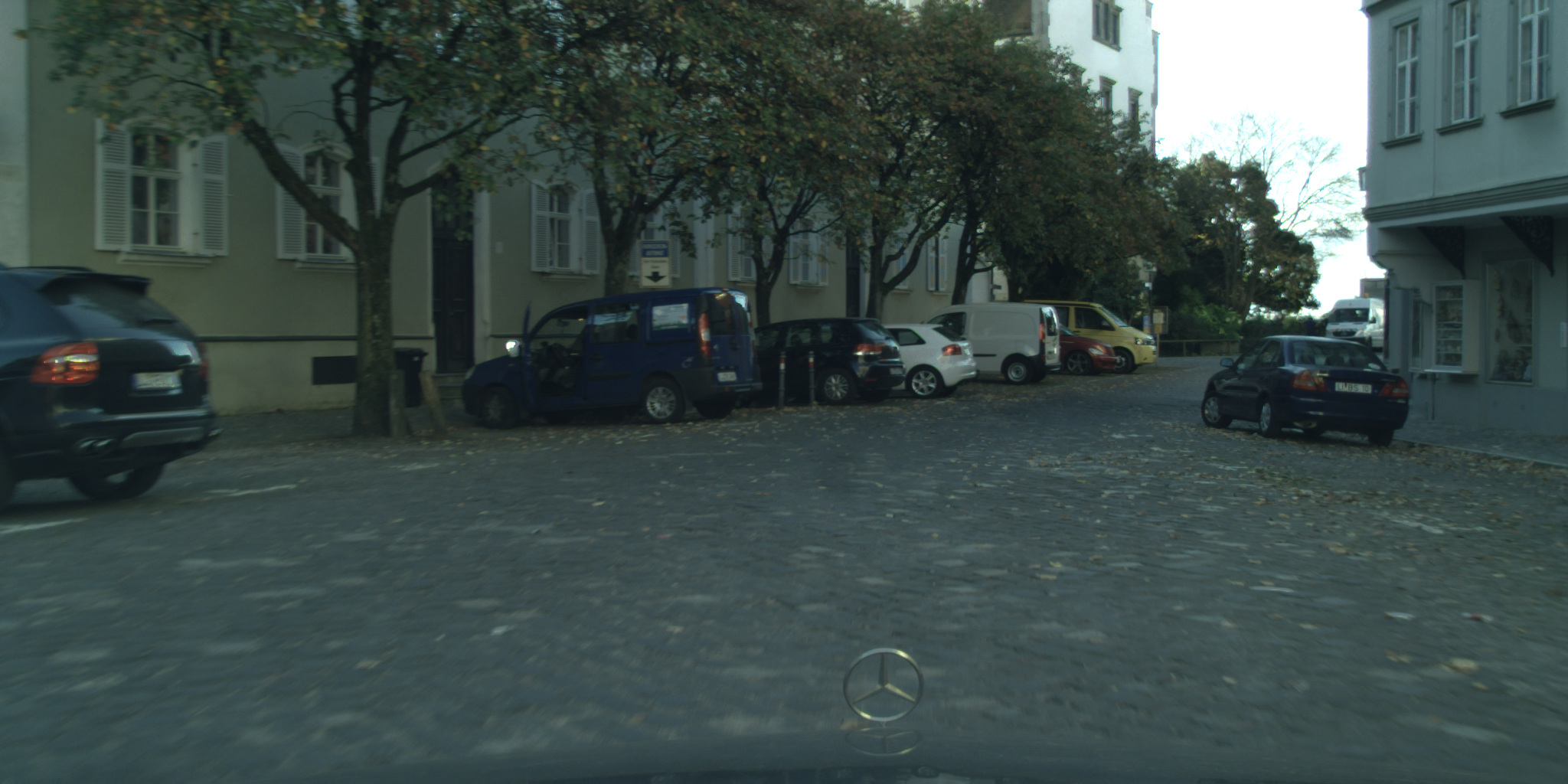}
    \end{minipage}%
 }%
  \subfigure[Ground Truth]{%
    \begin{minipage}[t]{0.2\linewidth}
      \centering
      \includegraphics[height=2cm,width=.9\linewidth]{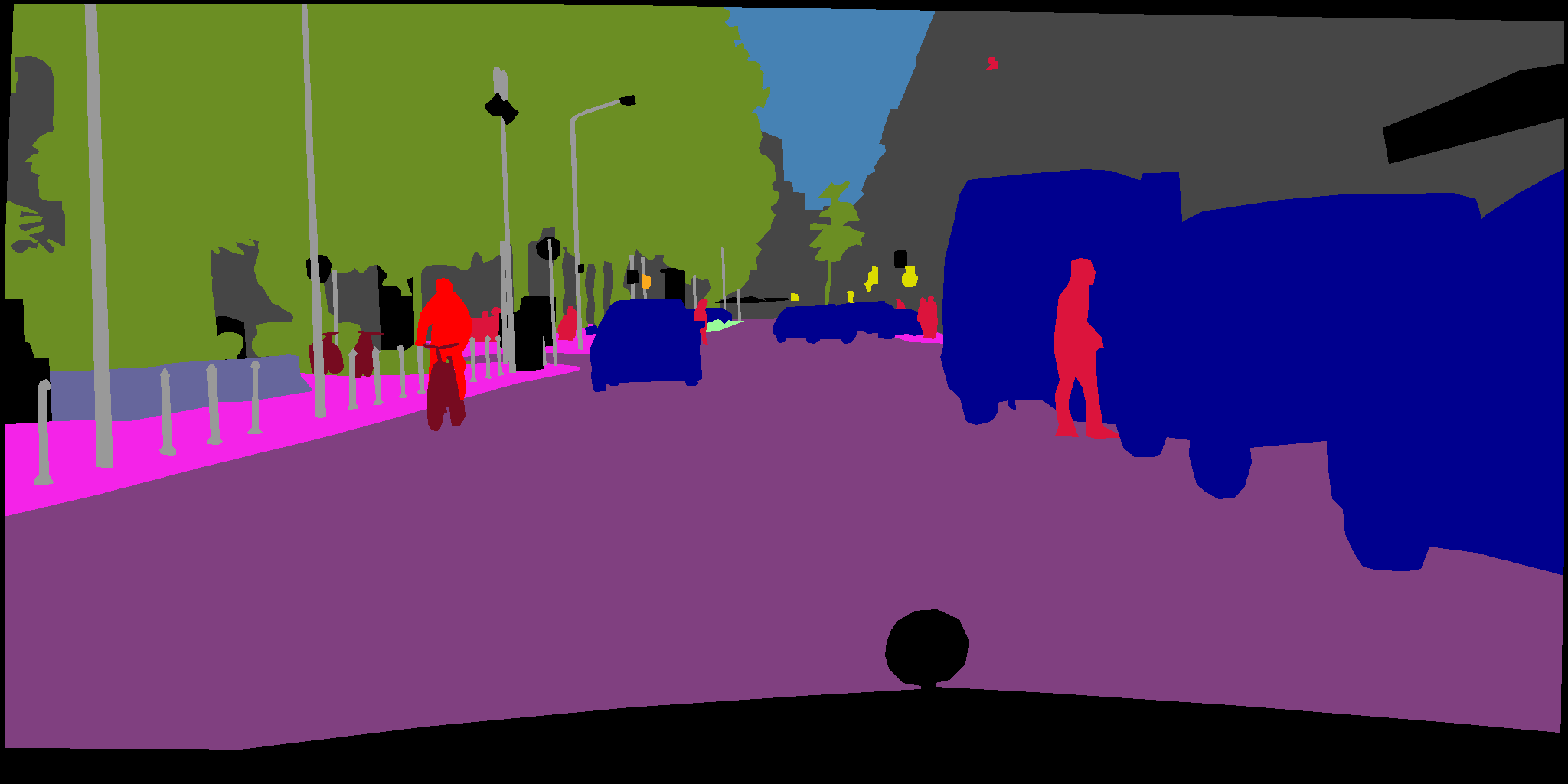}\\
      \vspace{4pt}
      \includegraphics[height=2cm,width=.9\linewidth]{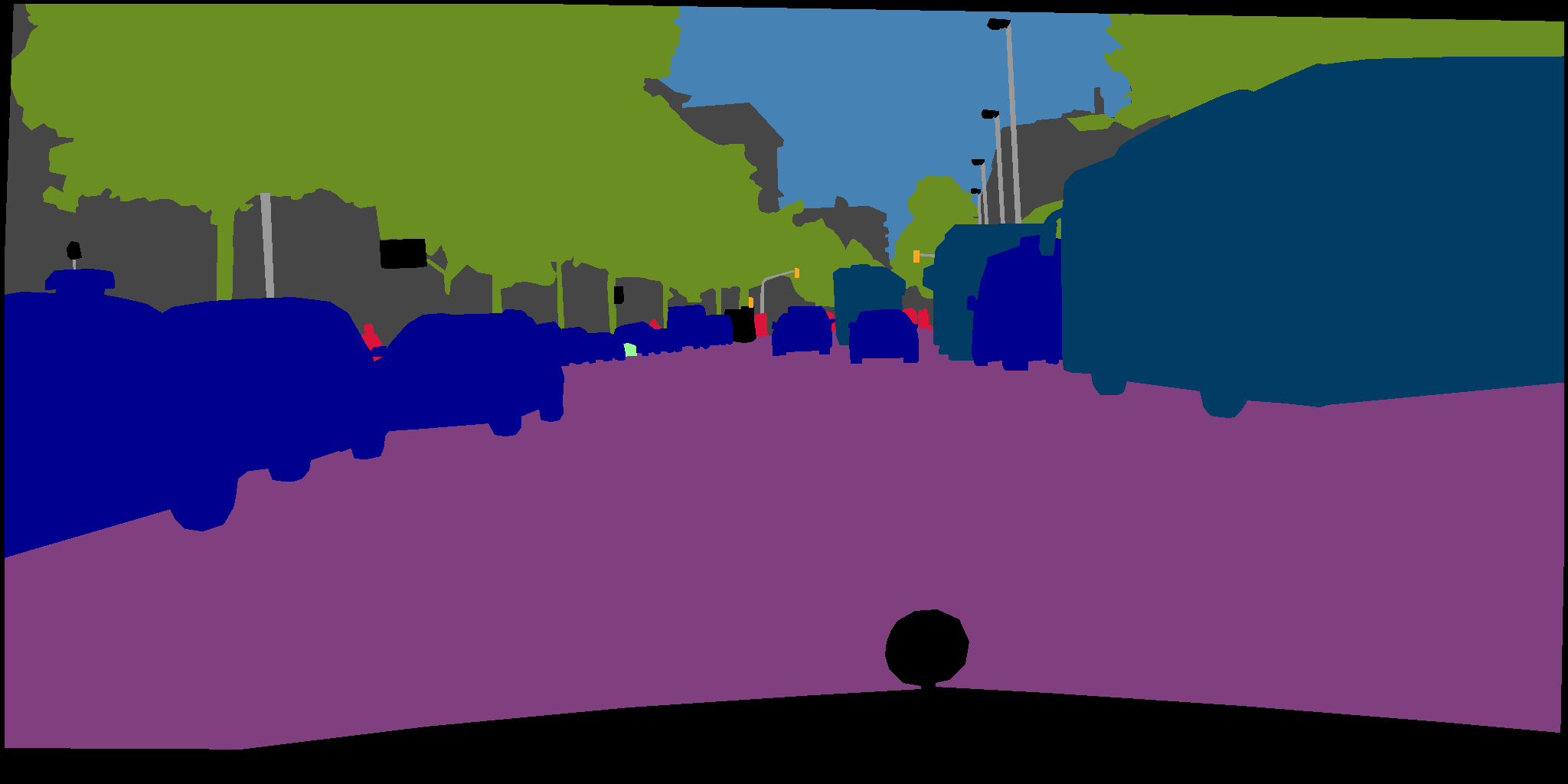}\\
      \vspace{4pt}
      \includegraphics[height=2cm,width=.9\linewidth]{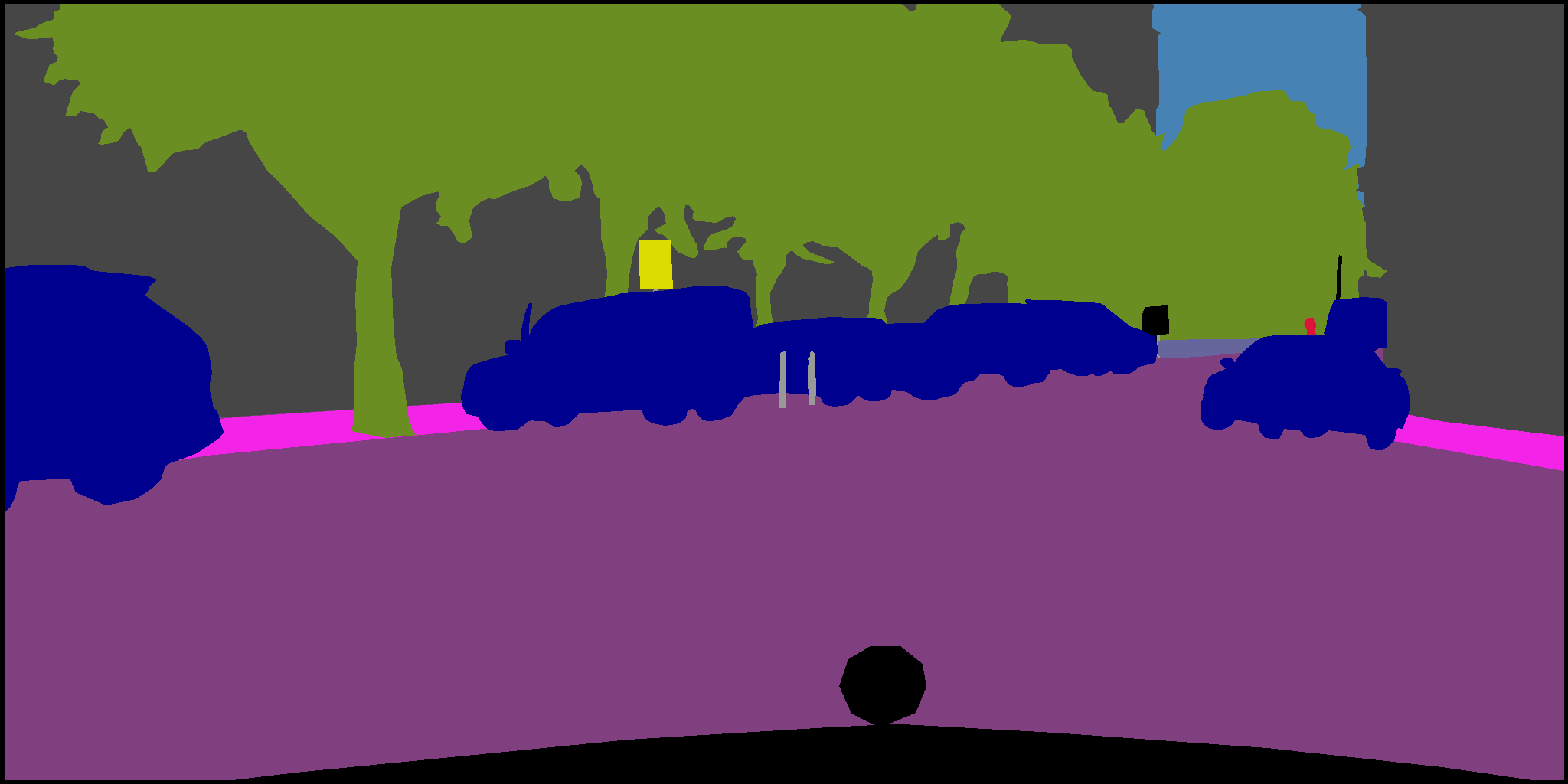}
    \end{minipage}%
  }%
  \subfigure[ENet]{%
    \begin{minipage}[t]{0.2\linewidth}
      \centering
      \includegraphics[height=2cm,width=.9\linewidth]{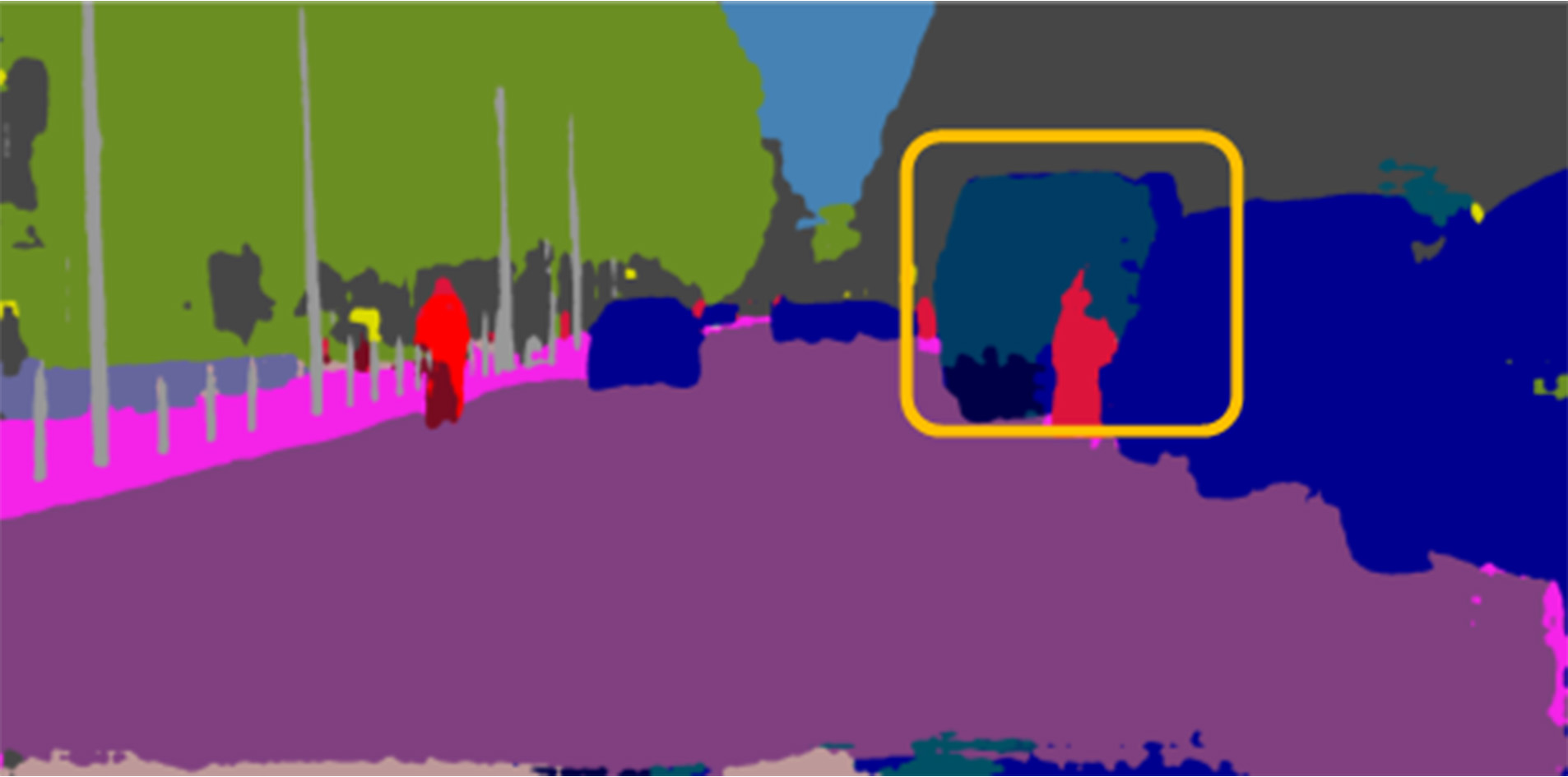}\\
      \vspace{4pt}
      \includegraphics[height=2cm,width=.9\linewidth]{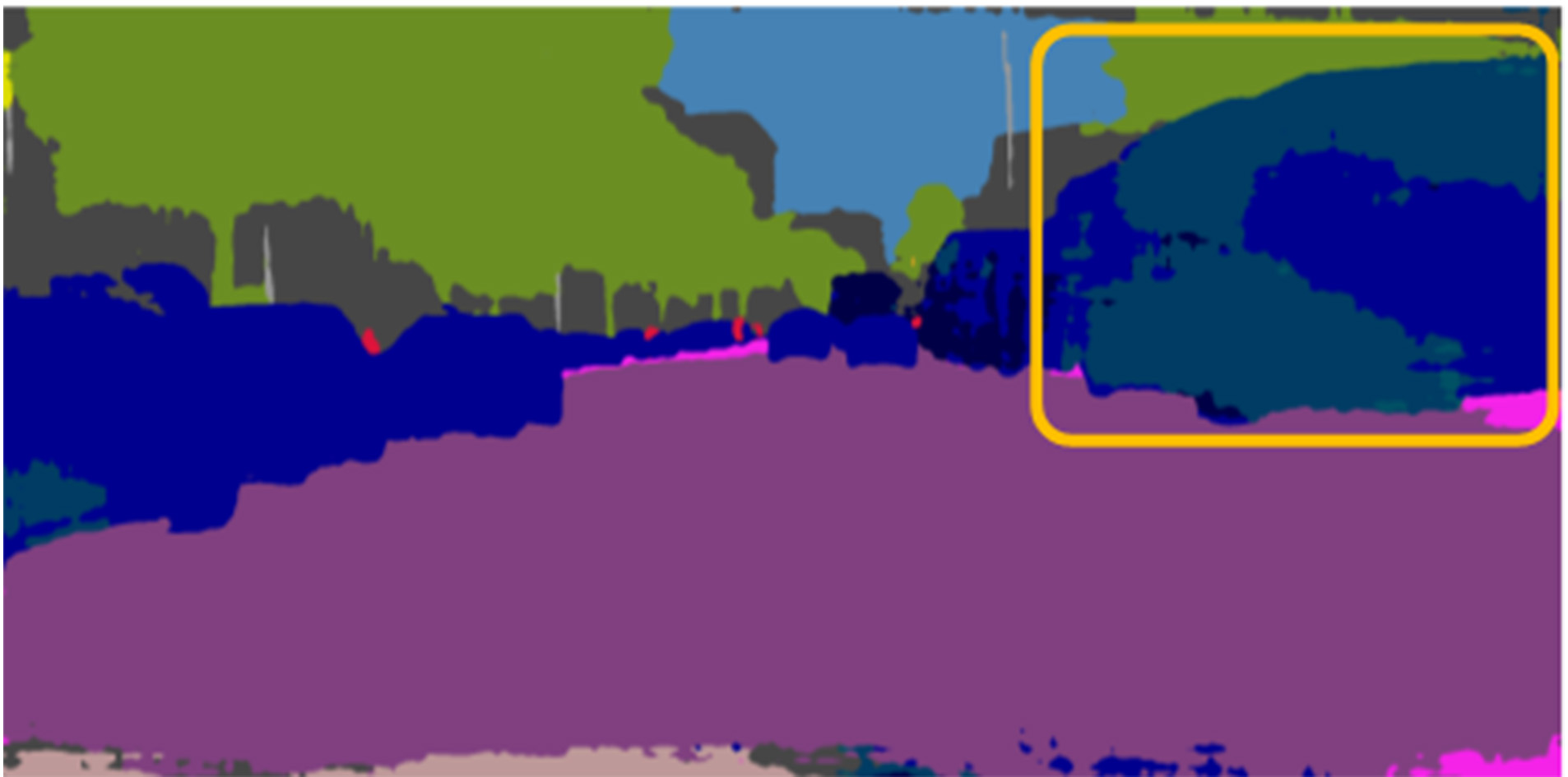}\\
      \vspace{4pt}
      \includegraphics[height=2cm,width=.9\linewidth]{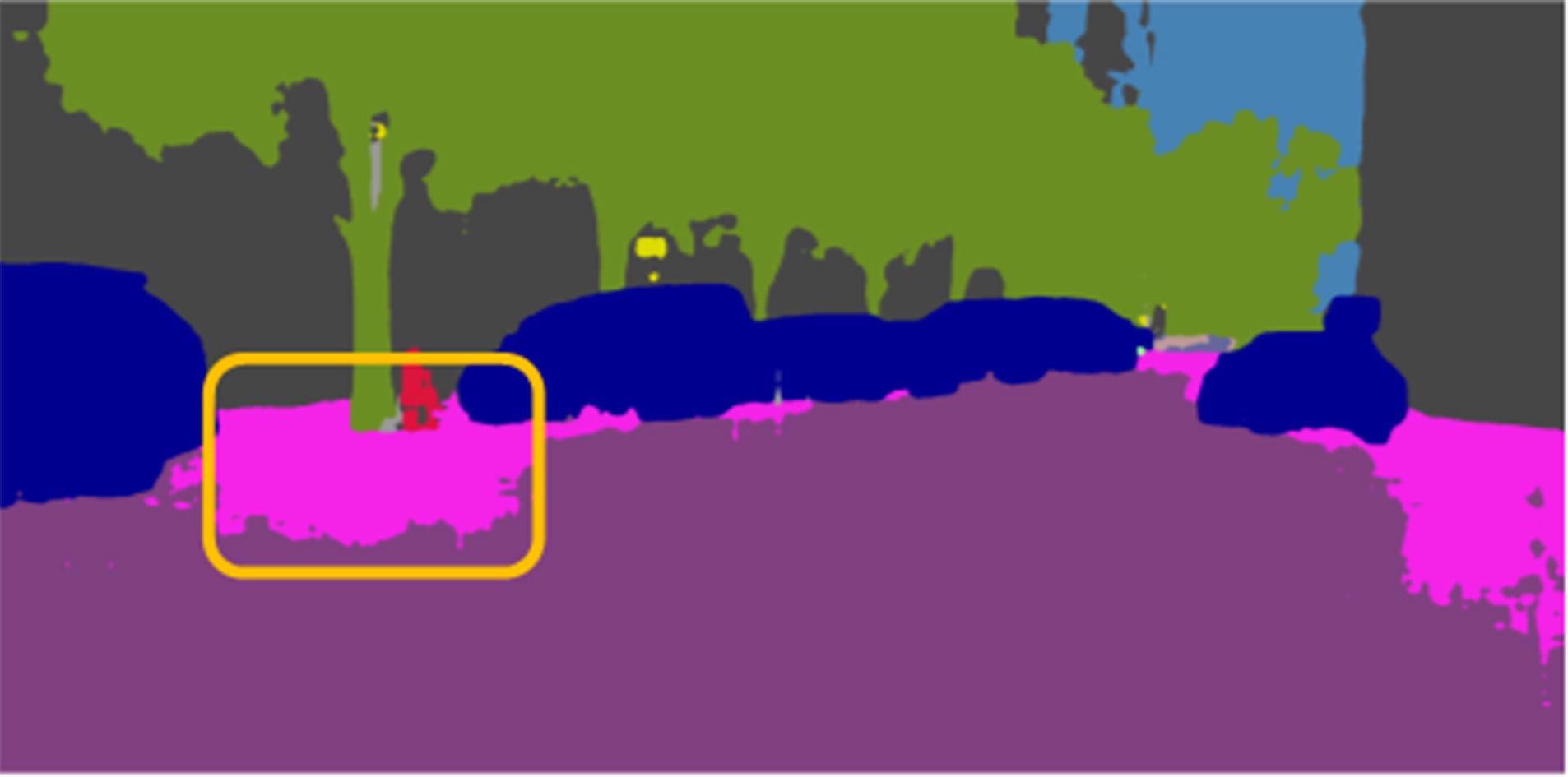}
    \end{minipage}%
  }%
  \subfigure[DABNet]{%
    \begin{minipage}[t]{0.2\linewidth}
      \centering
      \includegraphics[height=2cm,width=.9\linewidth]{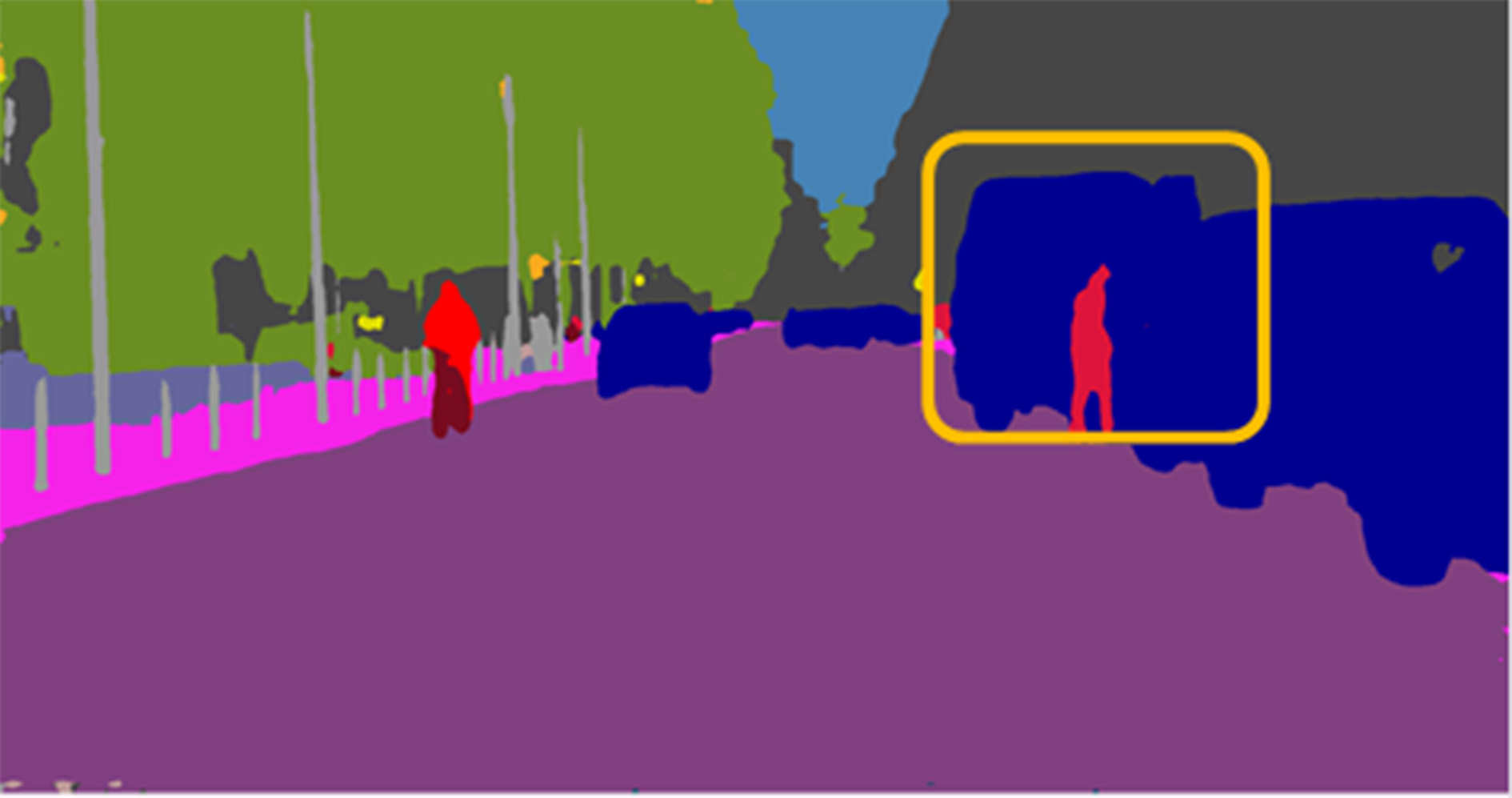}\\
      \vspace{4pt}
      \includegraphics[height=2cm,width=.9\linewidth]{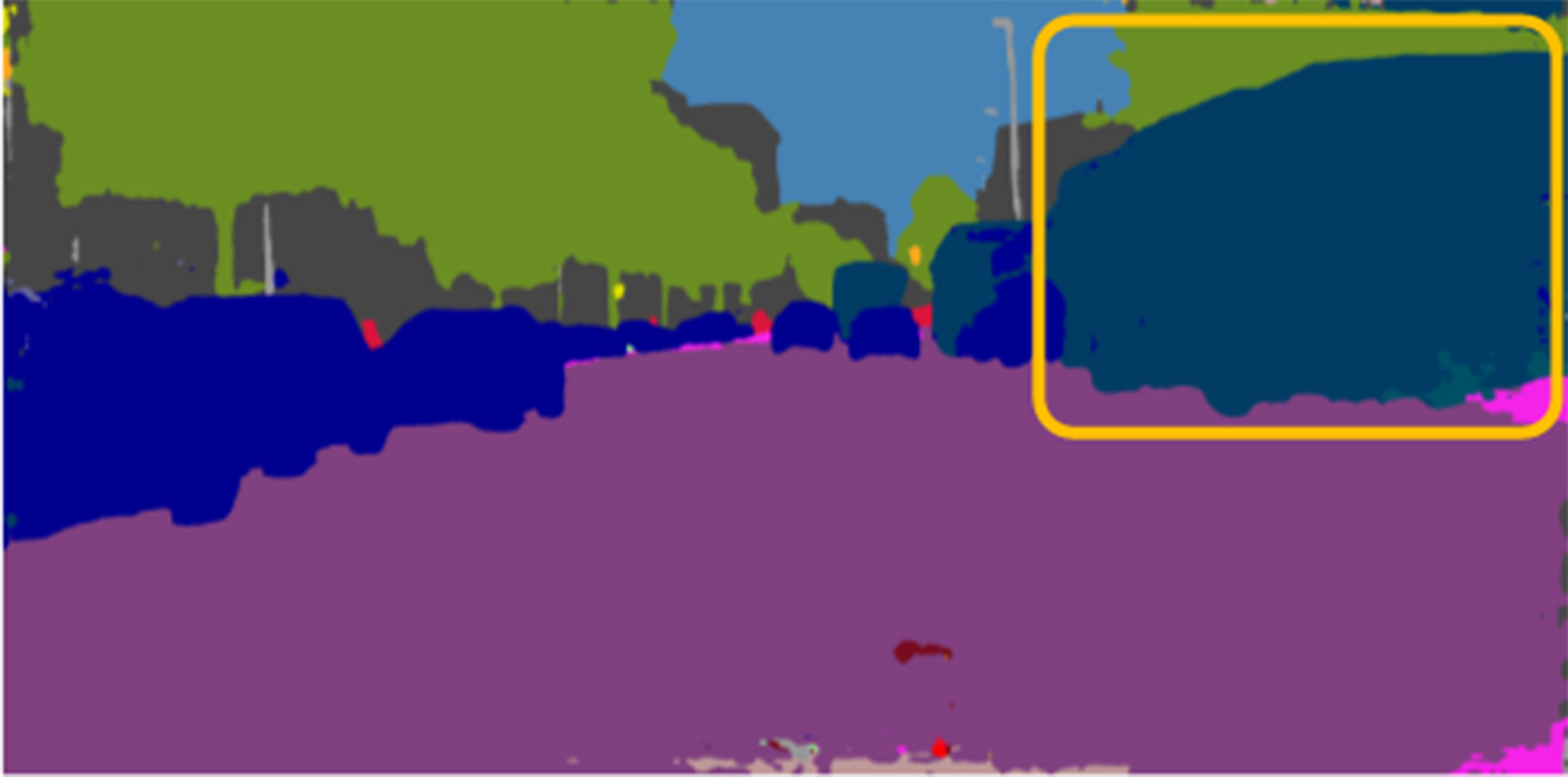}\\
      \vspace{4pt}
      \includegraphics[height=2cm,width=.9\linewidth]{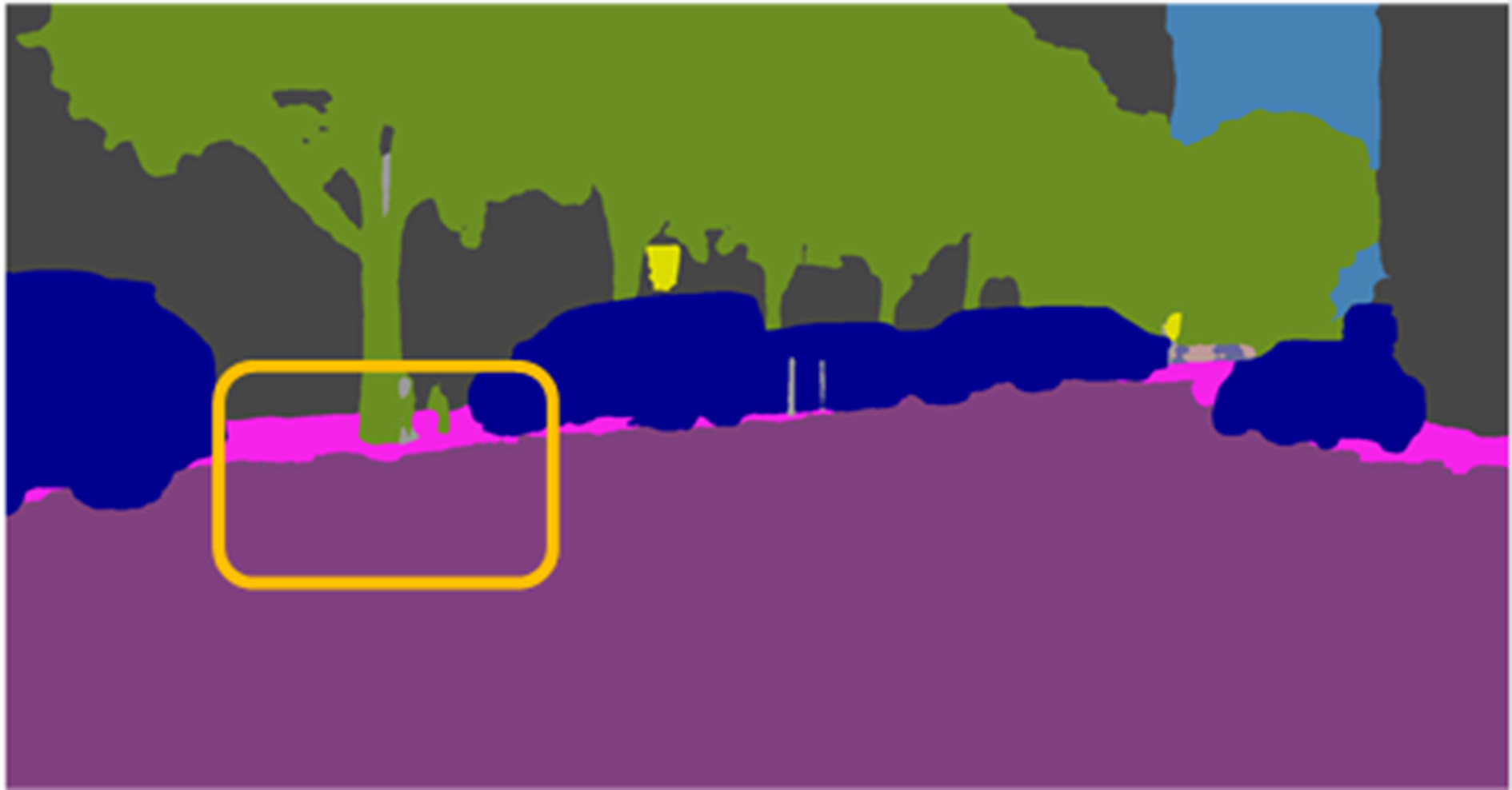}
    \end{minipage}%
  }%
  \subfigure[DPANet(ours)]{%
    \begin{minipage}[t]{0.2\linewidth}
      \centering
      \includegraphics[height=2cm,width=.9\linewidth]{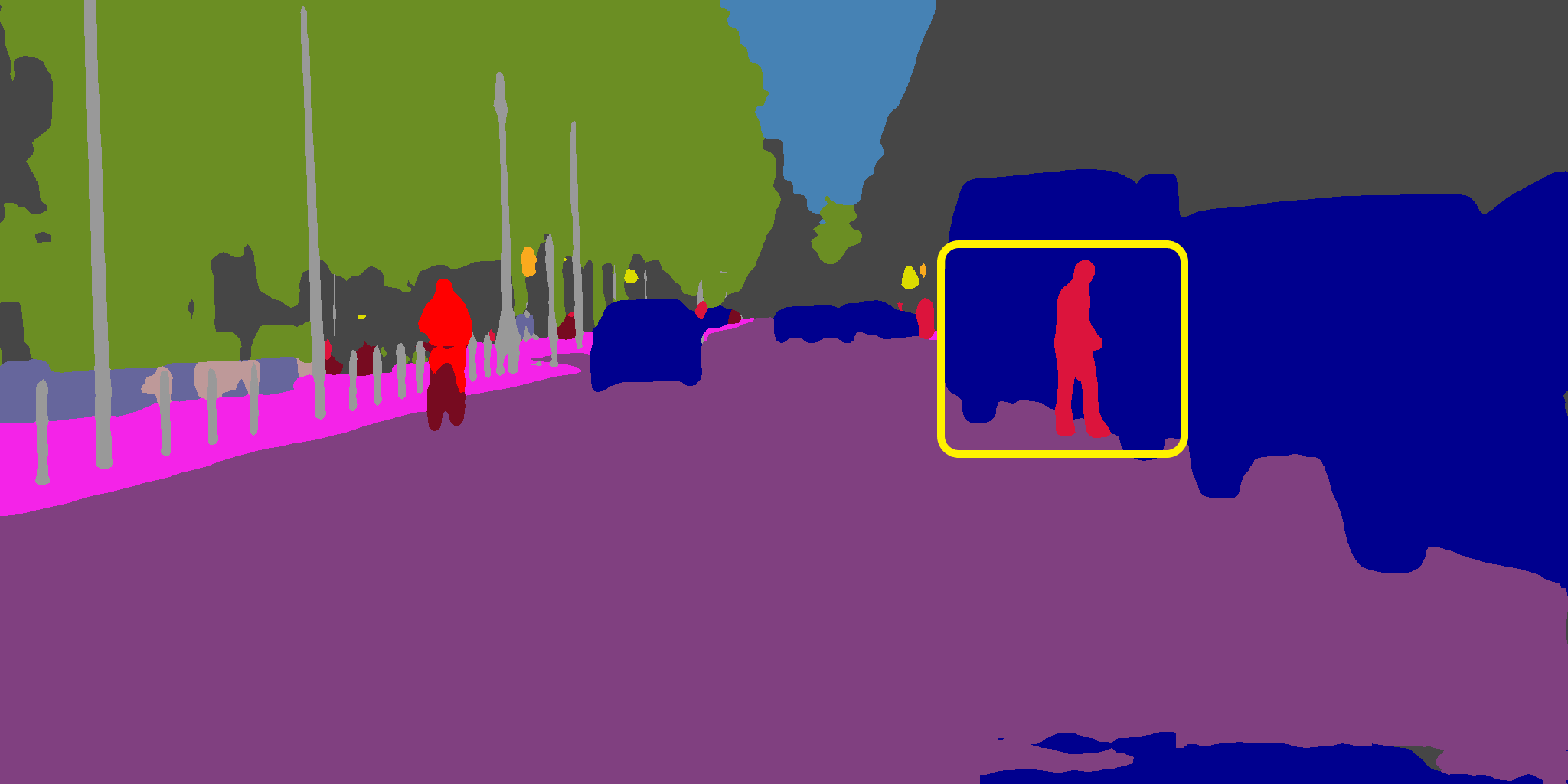}\\
      \vspace{4pt}
      \includegraphics[height=2cm,width=.9\linewidth]{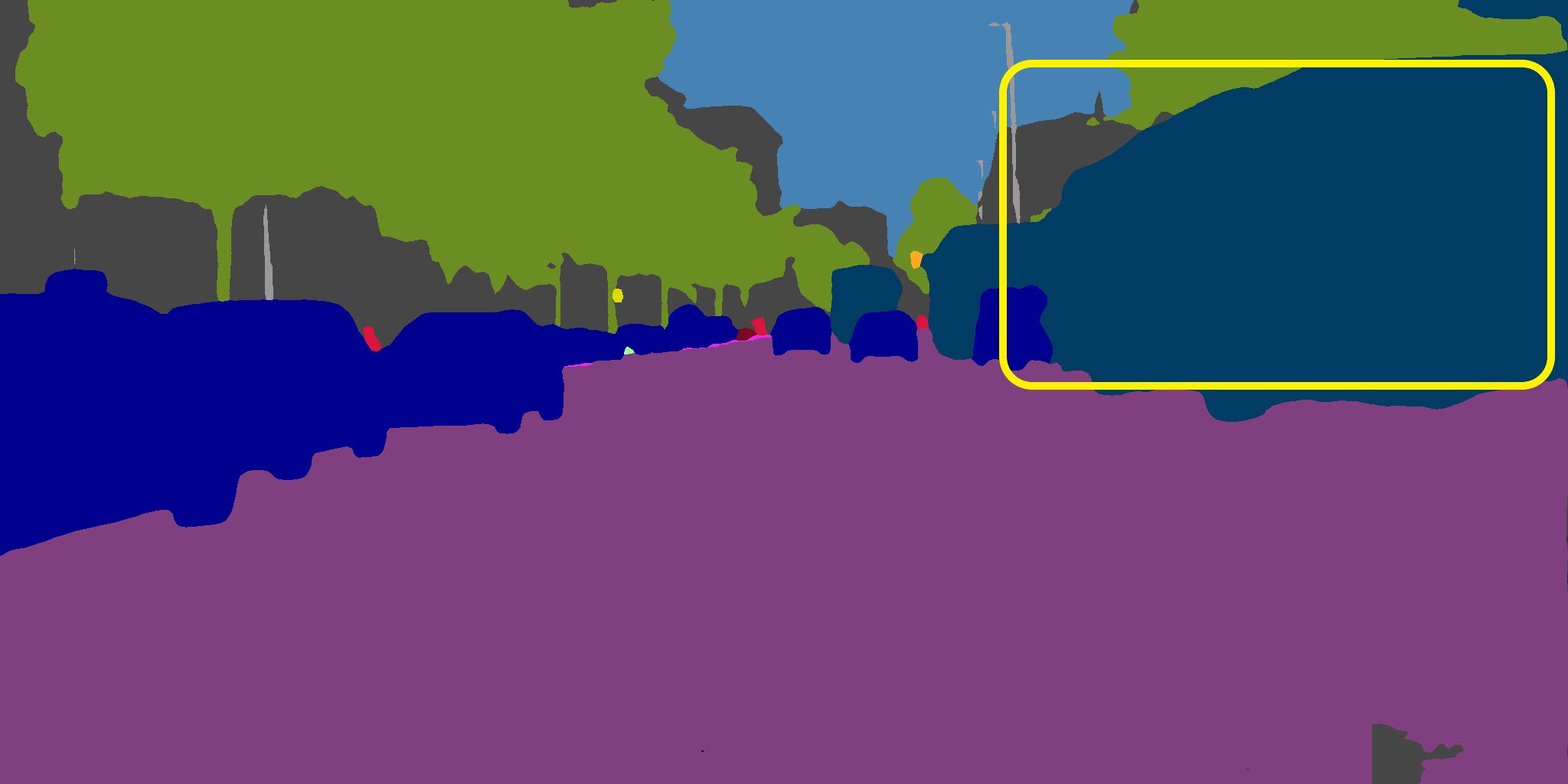}\\
      \vspace{4pt}
      \includegraphics[height=2cm,width=.9\linewidth]{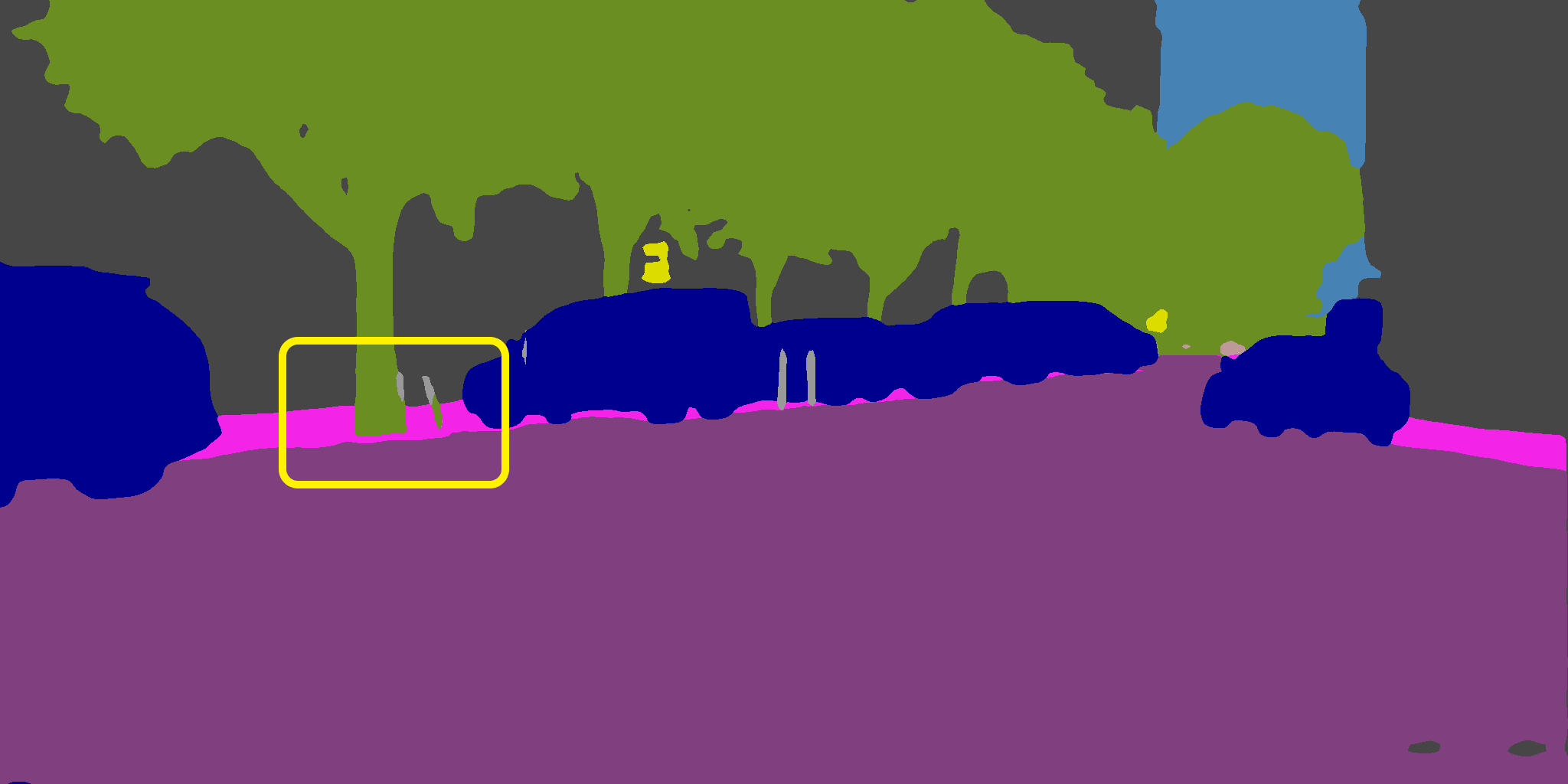}
    \end{minipage}%
  }%
	\caption{Qualitative comparisons with other methods.}
	\label{f22}
\end{figure}

We also give several typical comparison in Fig.\ref{f22}. It can be concluded from the observation that our proposed network gets a good segmentation effect on the person behind the truck and on the boundaries of the roads and plants.
\subsubsection{Ablation Study}
We adopt ablation study in following aspects: different backbones, combining different modules as illustrate in table \ref{tab:diff_backbone}. We embed ResNet50 or ResNet101 into network as backbone and whether use pre-train to verify our DPANet
effect.

\begin{table}[thb]\centering
    \caption{Different Backbone on the validation set of Cityscapes}
    \label{tab:diff_backbone}
    \resizebox{0.48\textwidth}{!}{
    \large
    \begin{tabular}{*{10}{c}}
        \toprule
       Method & BackBone &  PreTrained & SPM &  CPA & Mean IoU \% \\
        \midrule
        DPANet & ResNet50  &  &\checkmark  &\checkmark & 57.14\\
        DPANet & ResNet50  &\checkmark  &\checkmark  &\checkmark & 60.93\\
        \midrule
        DPANet & ResNet101 &  &\checkmark  &\checkmark & 63.912\\
        DPANet & ResNet101 &\checkmark  &\checkmark  &\checkmark & 69.46 \\
        \bottomrule
    \end{tabular}
    }
\end{table}

To further validate the effectiveness of our proposed modules in segmentation task, we adopt different combinations to compose new network to verify its effectiveness without pre-train as demonstrated in table\ref{tab:table3}. From the table, in terms of SPA module using mixed mode is better than using a single odd or even mode. As far as CAP module, though experiment we recommend using subtraction to measure relationship of all channels.

\subsubsection{Results on PASCAL VOC 2012}
We effectuate experiments on PASCAL VOC 2012 dataset.
In order to improve effectiveness of our DAPNet. The results are shown in Table.\ref{tab:PASCAL VOC 2012}. Our DPANet obtains a competitive results. Though the mean IoU of dilated FCN-2s network is higher than ours, we have a lower params and FLOPs.
\begin{table}[ht]\centering
    \caption{Ablation study on different modules}
    \label{tab:table3}
    \resizebox{0.48\textwidth}{!}
{
    \large
    \begin{tabular}{*{12}{c}}
        \toprule
        &  & \multicolumn{3}{c}{SPA Module} & \multicolumn{3}{c}{CPA Module} & \multicolumn{3}{c}{Mean IoU\%}\\
        Method & Backbone &  \makecell{Only Odd} &  \makecell{Only Even} &  \makecell{Mixed} &  \makecell{} &  \makecell{Subtract} & \makecell{Square} \\
        \midrule
        DPANet & ResNet101 &\checkmark  &  &  &  & \checkmark  & & 61.98\\
        DPANet & ResNet101 &  & \checkmark &  &  & \checkmark & & 61.87 \\
        DPANet & ResNet101 &  &  & \checkmark  &  &  & \checkmark  & 60.14\\
        DPANet & ResNet101 &  &  & \checkmark  &  & \checkmark &  & 62.0\\
        \bottomrule
    \end{tabular}
}
\end{table}

\begin{table}[ht]
    \caption{The results on PASCAL VOC 2012.}
    \label{tab:PASCAL VOC 2012}
    \centering
    \begin{tabular}{c|c|c}
      \midrule
      Method& Backbone & MIoU(\%)\\
      \midrule  
      FCN& vgg16& 62.2\\
      ESPNet\cite{mehta2018espnet}&No&63.0\\
      SSDD\cite{shimoda2019self}&FCN&65.5\\
      RRM\cite{zhang2020reliability}& No& 66.5\\
      FCN& No& 67.2\\
      Dilated FCN-2s VGG19\cite{kamran2018efficient}& VGG19& \textbf{69.0}\\
      \midrule
      DPANet&ResNest101&68.9\\
      \midrule
    \end{tabular}
  \end{table}
  
\newpage
\section{Conclusion}
In this paper, we analyse and study current situation of semantic segmentation with deep learning as the main method.Then, we designed a novel attention network called DPANet for semantic segmentation based on  prevalent attention mechanism. Our approach uses odd and even different scales pooling operations process feature maps to generate a attention mask which densely contains spatial characteristics but non-locally. For channel part, because holistic channels regarded as a distribution map of weight. Extending or compressing channels number is likely as a Fourier or Laplace transform, the core of that is different channels have different proportion contribution to an image. Thus, our channel attention module tries to figure out contribution level of channels. We also perform competitive results on segmentation datasets. We hope the designed network will offer help in transformer framework in future.



\bibliographystyle{unsrt}  
\bibliography{references}

\end{document}